\pdfoutput=1

\documentclass[11pt]{article}

\usepackage[preprint]{acl}

\usepackage{times}
\usepackage{latexsym}
\usepackage{xcolor} 
\usepackage{booktabs}
\usepackage[T1]{fontenc}

\usepackage[utf8]{inputenc}

\usepackage{microtype}

\usepackage{inconsolata}

\usepackage{graphicx}
\usepackage[most]{tcolorbox}

\title{Shakespearean Sparks: The Dance of Hallucination and Creativity in LLMs' Decoding Layers}

\author{Zicong He $\thanks{Equal Contribution}$ \\
  Georgia Institute of Technology \\
  \texttt{zhe384@gatech.edu} \\\And
  Boxuan Zhang $\footnotemark[1]$ \\
  Wuhan University \\
  \texttt{boxzhang1005@whu.edu.cn} \\\And
  Lu Cheng $\thanks{Correspondence to Lu Cheng (\href{lucheng@uic.edu}{lucheng@uic.edu})}$ \\
  University of Illinois Chicago \\
  \texttt{lucheng@uic.edu} \\}

\begin{document}
\maketitle
\begin{abstract}

Large language models (LLMs) are known to hallucinate, a phenomenon often linked to creativity. While previous research has primarily explored this connection through theoretical or qualitative lenses, our work takes a quantitative approach to systematically examine the relationship between hallucination and creativity in LLMs. Given the complex nature of creativity, we propose a narrow definition tailored to LLMs and introduce an evaluation framework, HCL, which quantifies \textbf{H}allucination and \textbf{C}reativity across different \textbf{L}ayers of LLMs during decoding. Our empirical analysis reveals a tradeoff between hallucination and creativity that is consistent across layer depth, model type, and model size. Notably, across different model architectures, we identify a specific layer at each model size that optimally balances this tradeoff. Additionally, the optimal layer tends to appear in the early layers of larger models, and the confidence of the model is also significantly higher at this layer. These findings provide a quantitative perspective that offers new insights into the interplay between LLM creativity and hallucination. The code and data for our experiments are available at \url{https://github.com/ZicongHe2002/HCL-Spark}.
\end{abstract}

\section{Introduction}
\label{sec: intro}
LLMs have demonstrated exceptional performance across various aspects, often rivaling or even surpassing those of humans \citep{luo2024large,trinh2024solving,OpenAI2024b}. 
Among these, \textit{creativity} is a highly recognized capability of LLM, which allows it to be used in a variety of domains, including text generation \citep{radford2019language}, reasoning \citep{brown2020language}, and image synthesis \citep{ramesh2021zero}. 
However, the enhanced \textit{creativity} usually comes with an increased propensity for \textit{hallucination} \citep{jiang2024survey}, i.e., generating misleading information and risky behaviors \citep{orgad2024llms}, which significantly hinders their application especially in high-stakes scenarios such as finance \citep{wu2023bloomberggpt} and healthcare \citep{singhal2025toward,zhou2025unveiling}. To address this concern, a considerable body of research has been dedicated to detecting \citep{farquhar2024detecting, manakul2023selfcheckgpt} and mitigating \citep{chuang2023dola, du2023improving, li2024inference} hallucinations.

\begin{figure}[t!]
  \centering
  \includegraphics[width=0.50\textwidth]{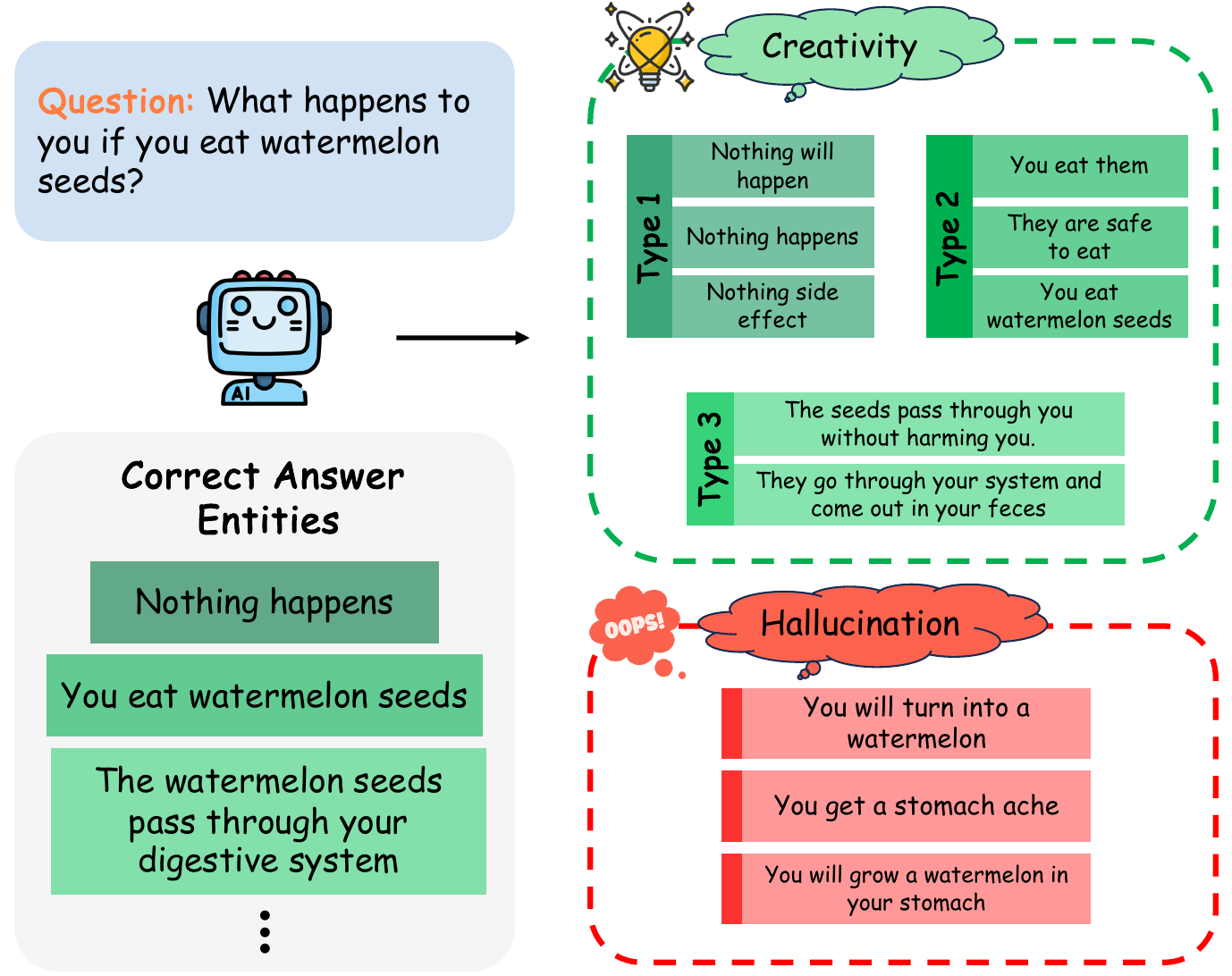}
  \caption{Illustration of our HCL evaluation criteria. Givena question with multiple correct answers, we instruct the LLM to generate various responses several times. Correct responses are shown in various shades of green, and creativity is defined as the diversity represented by distinct types grouped based on semantic similarities. Red boxes depict hallucinatory answers that are factually incorrect.}
   \vspace{-4mm}
  \label{fig:motivation}
\end{figure}


Recently, some efforts begin to delve into the connection between the two characteristics in LLMs \citep{lee2023mathematical, jiang2024survey}. 
From a philosophical perspective, as \textit{The Creativity Hidden in Hallucination} suggests, what is often dismissed as “wrong” may harbor unexpected creativity. For example, Copernicus’s heliocentric theory was initially regarded as heresy, yet it eventually revolutionized the field of astronomy \citep{jiang2024survey}.
Although promising progress has been achieved, existing studies are still limited in theoretically or qualitatively exploring the relationship between creativity and hallucination, lacking a empirical and systematic study of this connection in LLMs.  
Simultaneously, current efforts centered on creativity assessments primarily explore on specific tasks such as storytelling \citep{gomez2023confederacy}, poetry \citep{chakrabarty2024art}, and artistic ideation \citep{lu2024llm}, lacking a general and accurate definition and quantification method for the creativity tailored to LLMs. More specifically, traditional approaches typically rely on predefined criteria (e.g., originality, content fluency, and character similarity) or comparisons against other generations. 
However, the inherently stochastic (i.e., generations vary across instances) and unpredictable hallucinations (i.e., false or inaccurate information) of LLM outputs make it difficult for established methods to accurately measure the creative capabilities of LLMs.

To fill the above gaps, we propose a novel framework to conduct the first empirical analyses of the interplay between creativity and hallucinations from the inner structure of LLMs, i.e., layer to layer. We refer to this framework as HCL (\textbf{H}allucination and \textbf{C}reativity across \textbf{L}ayers). 
Since the outputs directly generated by the early layers of LLM are usually unstable or even invalid \citep{elhoushi2024layer}, we adopt the \textit{Layer-Skip} \citep{elhoushi2024layer} to ensure the generated content are consistently meaningful during layer-wise response sampling. 
Each response is then subjected to factual and diversity verification and categorized into two classes: creativity and hallucination.
Following prior works \citep{orgad2024llms}, the hallucination indicator is assigned with the error rates among the generated responses.
For the creativity metric, we provide a narrow definition tailored to the LLM that quantifying it as the diversity of correctness among sampled responses for each layer (Figure \ref{fig:motivation}).
We conduct extensive empirical analyses to examine their connections and identify a broadly consistent tradeoff between hallucination and creativity across different layer depths and sizes of LLMs.
The combination of these two dimensional metrics consequently yields a hallucination-creativity balanced (HCB) score for each layer, assisting in locating the optimal decoding layer for different model architectures that tend to produce accurate and varied outputs.
Our contributions are summarized as follows:
\begin{enumerate}
    \item Conceptually, we study a new perspective to explore LLMs' inner structure regarding the relationship between \textit{creativity} and \textit{hallucination} in LLMs during generating responses in common question-answering domains.
    \item Technically, we propose a new evaluation framework, namely, HCL, to analyze the layer-wise evolution of creativity and hallucination in LLM's responses and the trade-offs between the two concepts.
    \item Empirically, Our experiments show several inspiring findings, including the observation that creativity always comes with hallucination in LLMs. Furthermore, from the perspective of balancing creativity and hallucination, we find that relying on the final layer's output is not always optimal. Instead, early-exiting at intermediate layers yields better performance.
    

    
\end{enumerate}

\begin{figure*}[t!]
  \centering
  \includegraphics[width=0.98\textwidth]{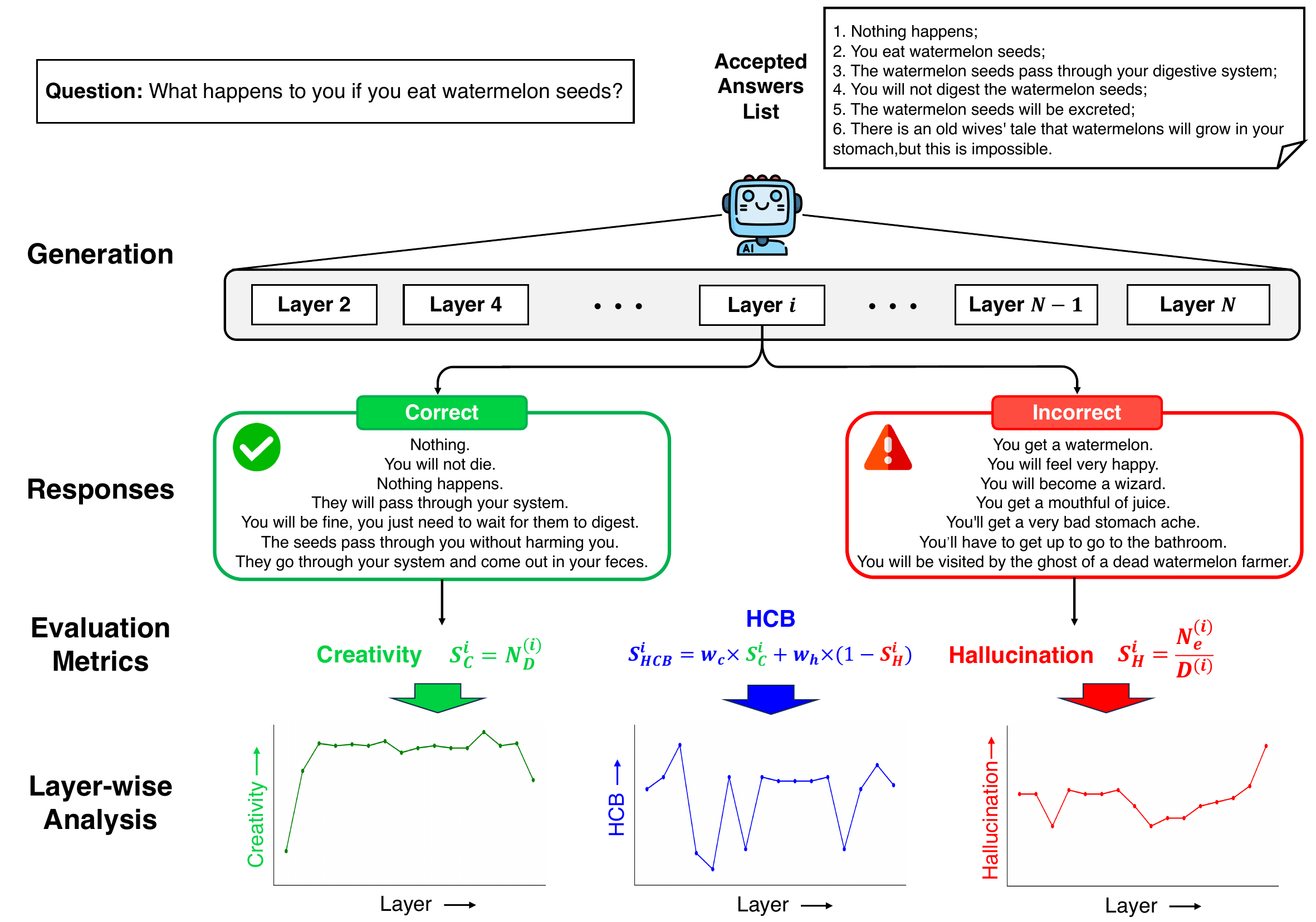}  
  \caption{Overview of our \textit{HCL} framework. We employ the \textit{layer\_skip} method, where each layer of the LLM is queried with the same prompt multiple times, generating diverse responses. The responses are then categorized into \textbf{correctness} and \textbf{hallucination}. Next, the correct responses undergo a secondary classification, where each color represents a distinct category of responses, collectively referred to as a type of \textbf{creativity}. Finally, we compute the \textbf{HCB score} by integrating the \textbf{creativity score} ($\mathbf{S_c}$) and the \textbf{hallucination score} ($\mathbf{S_H}$).}

  \label{fig:framework}
\end{figure*}

\section{Related Work}
\label{sec: related}
LLMs have demonstrated remarkable abilities in various domains, yet they still suffer from inherent issues such as hallucination and creativity uncertainty. While previous research has explored these two aspects separately, little attention has been given to their interplay. This section reviews existing work on hallucination and creativity in LLMs, highlighting the research gap that our study aims to address.

\paragraph{Hallucination in Large Language Models} 

Hallucination in LLMs refers to the generation of misleading, or incorrect content, which poses a significant challenge in high-stakes scenarios such as finance \citep{wu2023bloomberggpt} and healthcare \citep{singhal2025toward}. Extensive research has been conducted to detect and mitigate hallucinations in LLMs.
For hallucination detection, recent studies leverage self-verification mechanisms \citep{manakul2023selfcheckgpt}, confidence-based methods \citep{farquhar2024detecting}, and factuality assessments \citep{wang2024openfactcheck}. These approaches focus on identifying factually inconsistent outputs using external knowledge or entailment-based verification models.
For hallucination mitigation, methods such as Self-Reflection and Reasoning \citep{madaan2024self,mundler2023self,ji2023towards}, Prompt Tuning\citep{li2024inference,lester2021power,cheng2023uprise}, and retrieval-augmented generation (RAG) \citep{lewis2020retrieval,kang2023ever,gao2022rarr} have been proposed to improve factuality. However, these methods often lead to over-conservative generation, reducing the model’s ability to generate diverse and creative outputs.

While these approaches aim to eliminate hallucination, they do not consider its potential role in enhancing creativity. This raises the question of whether hallucination can contribute to novel and diverse responses, rather than being purely detrimental.

\paragraph{Creativity in Large Language Models}

Creativity in LLMs generally refers to their ability to generate novel, diverse, and contextually appropriate content. This capability has been widely applied in creative text generation.
Existing research primarily focuses on assessing and evaluating creativity in LLMs. As mentioned earlier, most studies assess LLMs’ creative potential by prompting them to generate content in domains such as storytelling \citep{gomez2023confederacy}, poetry generation \citep{chakrabarty2024art}, and artistic ideation \citep{lu2024llm}.The generated content is then evaluated using another, often superior, model that scores various aspects of creativity, such as originality, narrative fluency, flexibility, and refinement. This approach is commonly used to quantify the creative capabilities of LLMs.

Additionally, previous studies have conducted a mathematical analysis of the inherent trade-off between creativity and hallucination in LLMs and have demonstrated that hallucination is an intrinsic property of LLMs that, to some extent, enhances their creative potential\citep{lee2023mathematical}.This finding suggests that current creativity evaluation methods primarily focus on originality and coherence, potentially overlooking the role of hallucination in fostering creativity.

Despite the growing evidence revealing the inherent trade-off between hallucination and creativity \citep{jiang2024survey}, existing research still tends to treat them as independent phenomena. Most studies focus on reducing hallucination as an undesirable effect, while creativity research rarely considers the potential role of hallucination in generating innovative content. 

Therefore, at present, there is no systematic study investigating the relationship between hallucination and creativity in LLMs. This work aims to bridge this research gap.





\section{Methodology}
\label{sec:method}
In this study, we propose a three-stage evaluation framework  \textit{HCL (Hallucination-Creativity Layerwise)} to explore the relationship between creativity and hallucination in LLMs layer-wise generations (Figure \ref{fig:framework}). First, to ensure the layer-wise output is generally meaningful, we obtain the responses sampled from each layer of LLM by leveraging the early-exit strategy (Section \ref{sub:sec:sample}). Second, we propose the creativity metric and assign each response with both creativity and hallucination metrics (Section \ref{sub:sec:metric}). Lastly, we propose the HCB score which will be used to optimize the trade-off between these creativity and hallucination metrics (Section \ref{sub:sec:HCB}).

\subsection{Layer-wise Response Sampling}
\label{sub:sec:sample}

\begin{figure}[t!]
  \centering
  \includegraphics[width=0.48\textwidth]{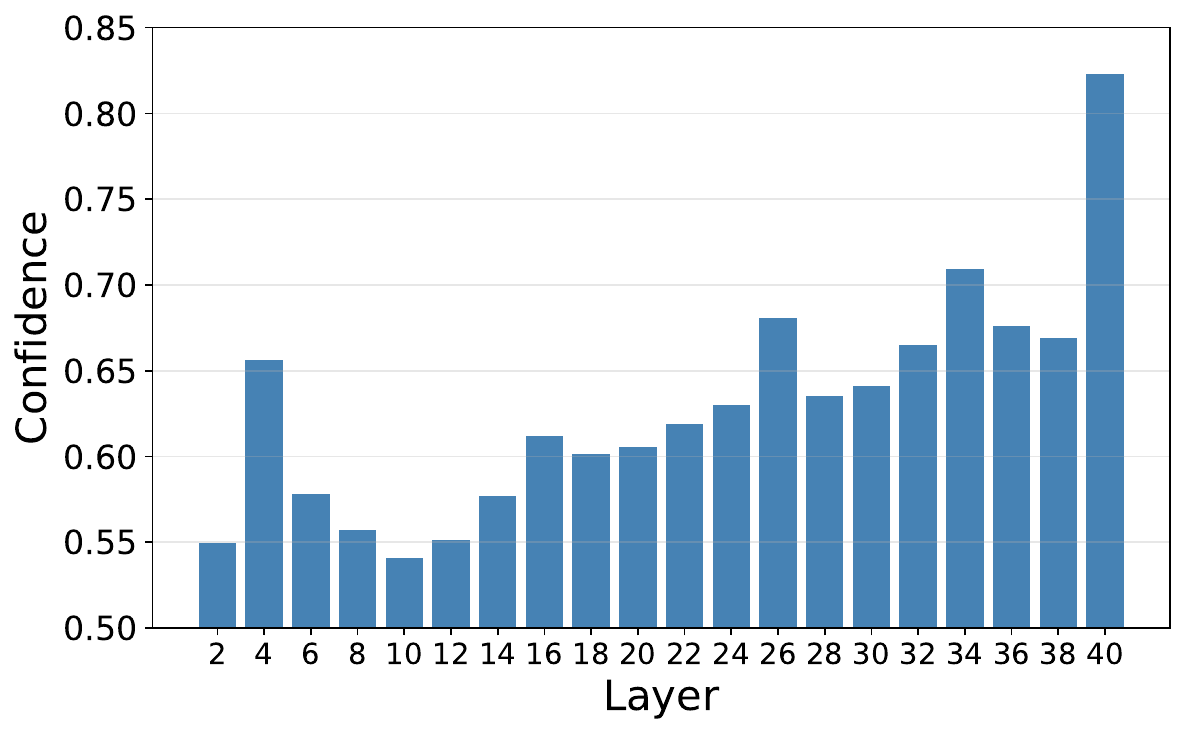}  
  \caption{Confidence variations across layers in LLaMA2-13B. We adopt P(True) to allow each layer of the LLM to self-evaluate the average confidence among the corresponding sampled responses.}
  \label{fig:confidence-13b}
\end{figure}

Unlike conventional decoding strategies that rely on the final layer's outputs, 
our key insight lies in \emph{analyzing and potentially utilizing the responses from intermediate layers}. 
This design is based on the following key observations and findings:

\begin{itemize}
    \item \textbf{Confidence is lower in earlier layers, enabling more diverse outputs.}
    During the decoding process of LLMs, earlier layers tend to exhibit higher uncertainty, preserving more possibilities in the generation process, as shown in Figure \ref{fig:confidence-13b}. This uncertainty allows them to produce more diverse and creative outputs. Furthermore, if these earlier layers can generate creative content with minimal impact on accuracy, it becomes feasible to directly extract responses from them improve the inference efficiency.

    \textbf{The need for early exit.}
    Since deeper layers tend to produce more conservative outputs, 
    while some intermediate layers may already achieve an optimal balance between creativity and hallucination, terminating decoding at these layers can not only reduce computational overhead but also prevent creativity loss ~\citep{chuang2023dola}.
\end{itemize}

Based on these observations and assumptions, we aim to \textbf{analyze creativity and hallucination layer by layer} to achieve two objectives:  
(1) Conduct a more fine-grained investigation into their interaction during the response generation process of LLMs, unveiling their underlying mechanisms.  
(2) Identify the optimal decoding layer that allows the model to exit early while maintaining a favorable balance between creativity and factual accuracy, thereby reducing computational cost.

In order to better understand how creativity and hallucination evolve across different depths,
we adopt a \emph{Layer-Skip} strategy inspired by speculative decoding~\citep{elhoushi2024layer}. Specifically, given an input consisting of a question $q$ and a shared prompt $p$,
we sample responses generated from the earlier layers $\{\ell_1, \ell_2, \dots, \ell_{N-1}\}$ (using speculative decoding) 
and the final layer $\ell_N$ (using standard autoregressive decoding) of the LLM.
We denote the resulting response list as $r$, formally expressed as:
\begin{equation}
\begin{aligned}
& r = \{[r_1, r_2, \dots, r_{N-1}], r_N\}, \\
\text{where} \ r_i & = \bigcup_{j=1}^D LLM_{i}^{(j)}(p(q)), \ i \in \{1, \dots, N\}.
\end{aligned}
\end{equation}

where $i$ refers to the i-th layer of the LLM and $D$ denotes the sampling times. Building upon the above procedure, we assigned $N \times D$ responses generated by each layer of the LLM to each question for subsequent layer-wise evaluation of the two metrics, creativity and hallucination.

\subsection{Evaluation Metric}
\label{sub:sec:metric}


\paragraph{Hallucination.}
Following \citep{orgad2024llms}, we define hallucination as any type of error generated by an LLM in our study. Hence, we have to justify the correctness of the responses generated by each decoding layer from LLM before evaluating their hallucination metrics. 
We adopt the following criteria for judging the correctness of free-form responses: if the generated response contains the correct answer, it is deemed correct; otherwise deemed hallucination.
Based on the above, the hallucination metric of sampled layer-wise responses can be defined as follows,

\begin{equation}
\begin{aligned}
\mathbf{S_H^i} = \frac{N_e^{(i)}}{D^{(i)}}, 
\quad \text{where} \ i \in \{1, \dots, N\}.
\end{aligned}
\end{equation}
where \(N_e^{(i)}\) denotes the incorrect times and \(D^{(i)}\) refers to the sampling time at \textit{layer $i$}.
\paragraph{Creativity.}


Following the definition that creativity is both novel and useful \citep{jiang2024survey}, we define the diversity of correct outputs as creativity. Therefore, when we filter out incorrect responses from the $n$ responses, we need to group the semantically equivalent \citep{ribeiro2018semantically} correct responses. To meet this requirement, we utilize a SentenceTransformer-based encoder, 
the pre-trained \textit{all-MiniLM-L6-v2} model \citep{vergou2023readability}, to extract dense semantic embeddings and group them as different semantic clusters based on the semantic-level similarity. Subsequently, we categorize the outputs based on group types and evaluate the creativity metric.
\begin{equation}
    \mathbf{S_C^i} = {N_D^{(i)}},
    \quad \text{where} \ i \in \{1, \dots, N\}.
\end{equation}
where $N_D^{(i)}$ is the count of unique semantic clusters at \textit{layer $i$}. 

\subsection{HCB Calculation}
\label{sub:sec:HCB}

Once we obtain the scores for \textbf{creativity} and \textbf{hallucination}, we need to evaluate the performance of each model layer in generation tasks. To achieve this, we propose a \emph{Hallucination and Creativity Balanced (HCB)} Score, which combines \textbf{creativity} and \textbf{hallucination} using distinct normalization methods. Specifically, \textbf{creativity} is normalized via min-max scaling, while \textbf{hallucination} is quantified directly through the error rate. This score provides a unified metric to assess the model's ability to generate outputs that are both accurate and diverse, ensuring a balanced trade-off between creativity and hallucination.

We compute the HCB score \(S_{HCB}^i\)in the layer $i$ as follows:

\[
S_{HCB}^i = w_c \times S_C^i \;+\; w_h \times \bigl(1 - S_H^i\bigr),
\]

where \(w_c\) and \(w_h\) are the weights corresponding to creativity and hallucination, respectively. Here, \(w_c + w_h = 1\). Note that \(S_C^i\) is the normalized score, where \(S_C^i\) is the normalized creativity score, and \(S_H^i\) is the hallucination score, and \(S_{HCB}^i\) is the HCB score.

\section{Experiments}
\label{sec: exp}
In this section, we present the experimental setup, models, datasets, and discuss the key findings. Based on previous methods (Section \ref{sub:sec:HCB}), in all experiments, for each query, LLMs respond 50 times using the same prompt to ensure we have sufficient responses to evaluate the creativity and hallucination of LLMs.
\subsection{Experimental Setups}
\paragraph{Models}
\label{sec: models}


We use four popular open-weight models: LLaMA 3.2-1B, LLaMA 2-7B, LLaMA 3-8B, and LLaMA 2-13B~\citep{touvron2023LLaMA}. These models allow us to systematically analyze how model size and different layers influence the trade-off between creativity and hallucination.


\begin{figure*}[t!]
  \centering
  \includegraphics[width=1\textwidth]{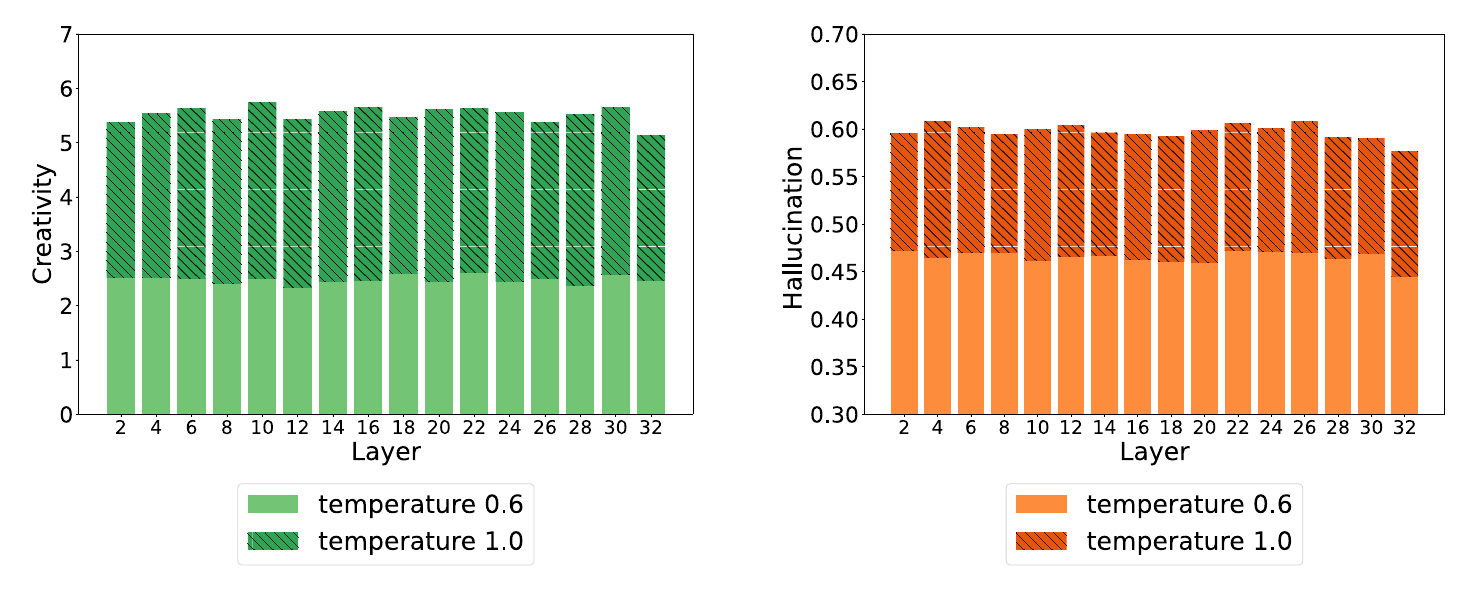}  
  \vspace{-6mm}
  \caption{The variation of layer-wise creativity and hallucination metrics of the LLaMA3-8B when its temperature coefficient increases from 0.6 to 1.0 on TriviaQA benchmark.}
  \label{fig:finding1-1}
  \vspace{-2mm}
\end{figure*}

\paragraph{Datasets}
\label{sec:datasets}

For our experiments, we utilized two open-domain question answering (QA) datasets: TriviaQA\citep{joshi2017triviaqa} and Natural Questions (NQ)\citep{kwiatkowski2019natural}. These datasets are widely used in QA research, covering a vast range of real-world questions with multiple valid answers. They provide a suitable benchmark for evaluating LLMs in terms of information retrieval, factual generation, and creative expression.

\textit{TriviaQA: }  
TriviaQA is a general knowledge QA dataset that spans multiple domains, including history, science, literature, sports, and entertainment. One of its key characteristics is that each question typically has multiple acceptable correct answers. This feature makes it ideal for assessing both the accuracy and creativity of LLMs, allowing evaluation even when models generate different but reasonable responses.

\textit{Natural Questions (NQ): }  
Natural Questions, released by Google, consists of real user queries from Google Search, with answers typically extracted from Wikipedia, emphasizing factual consistency. In the latest version of the dataset, Natural Questions have evolved from multiple-choice to open-ended text generation, introducing more flexibility. Moreover, the dataset now include many questions with multiple valid answers, making it more suitable for assessing response diversity.

In this study, we specifically filtered questions with three or more correct answers to ensure sufficient answer diversity. This approach allows us to assess whether models can maintain factual accuracy while exhibiting creativity, providing a more comprehensive evaluation of LLM performance in open-domain QA tasks.

\subsection{Explore the relationship between creativity and hallucination}
In this part, we focus on analyzing the creativity and hallucination metrics of LLMs at each layer during response generation. Our experimental results reveal some fundamental relationships between the two dimensions, providing deeper insights into their interplay.
\begin{figure*}[t!]
  \centering
  \includegraphics[width=1.0\textwidth]{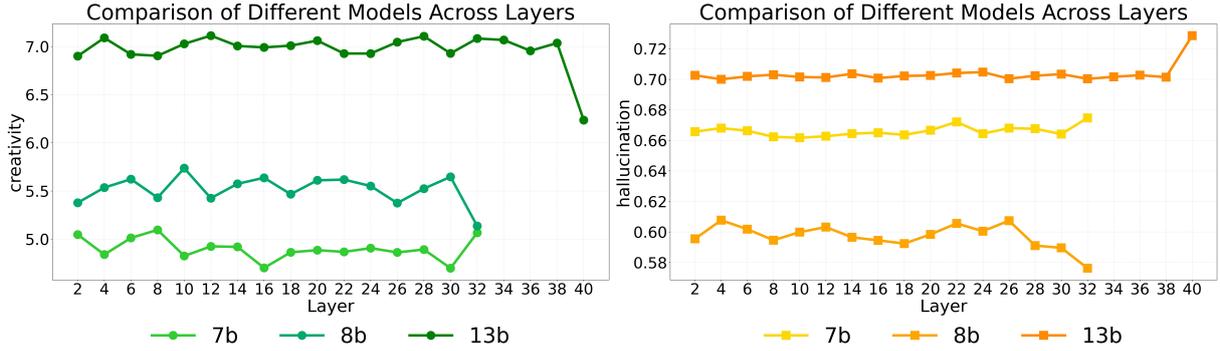}  
  \vspace{-6mm}
  \caption{The left figure illustrates the creativity scores across different models, while the right figure presents  the hallucination levels for the same models. Both evaluations were conducted with a temperature setting of 1.0. As observed, the LLaMA 2-13B model exhibits the highest creativity among all models. However, this increase in creativity also corresponds to a higher level of hallucination.
  }
  \vspace{-2mm}
  \label{fig:data_compare}
\end{figure*}
\paragraph{Creativity comes with hallucination. } 
Existing studies mainly consider increasing the model temperature to enhance the diversity of LLM's generations, since the temperature parameter determines the smoothness of the probabilities while sampling and a higher temperature value indicates more diverse sampling\citep{peeperkorn2024temperature}.
However, as the temperature parameter increases, both creativity and hallucination rates rise in a proportional manner, as shown in Figure \ref{fig:finding1-1}. This suggests that higher temperature values encourage more diverse and novel outputs, fostering greater creativity by allowing the model to explore unconventional ideas. However, this exploratory behavior comes at a cost: an increased likelihood of generating factually inaccurate or unverifiable content.

This trade-off highlights the inherent tension between diversity-driven creativity and factual precision in LLMs. When the model is set to lower temperatures, it tends to produce more deterministic and factually consistent responses, but at the expense of originality and expressiveness. Conversely, when the temperature is raised, the model exhibits a greater degree of unpredictability, leading to more imaginative but less reliable outputs. 

These findings align with previous studies, suggesting that LLMs’ tendency to hallucinate is not just a flaw but a natural consequence of their generative flexibility.

\paragraph{Stronger models are more creative, though also more prone to hallucination.}

A second key observation from our experiments is that LLMs tend to exhibit higher levels of both creativity and hallucination. Specifically, model size appears to correlate positively with the generation of novel yet sometimes factually incorrect responses. For instance, smaller models such as LLaMA-3.2-1B tend to be more conservative in their outputs, often adhering closely to more predictable, template-like responses. While this makes them less prone to hallucination, it also limits their ability to produce highly original and imaginative content. In contrast, larger models (e.g., LLaMA-3-8B or LLaMA-13B) demonstrate a greater ability to generate complex and creative responses, but they are also more susceptible to producing hallucination(Figure \ref{fig:data_compare}). This suggests an intrinsic trade-off between model capacity and output reliability: as models become more expressive and generative, they also gain a higher degree of unpredictability, leading to a higher risk of fabricating details that deviate from factual correctness. 

These findings underscore the dual-edged nature of language models. While larger models unlock greater generative potential, they require more robust control mechanisms to mitigate hallucinations.

\begin{figure}[t]
  \centering
  \includegraphics[width=0.48\textwidth]{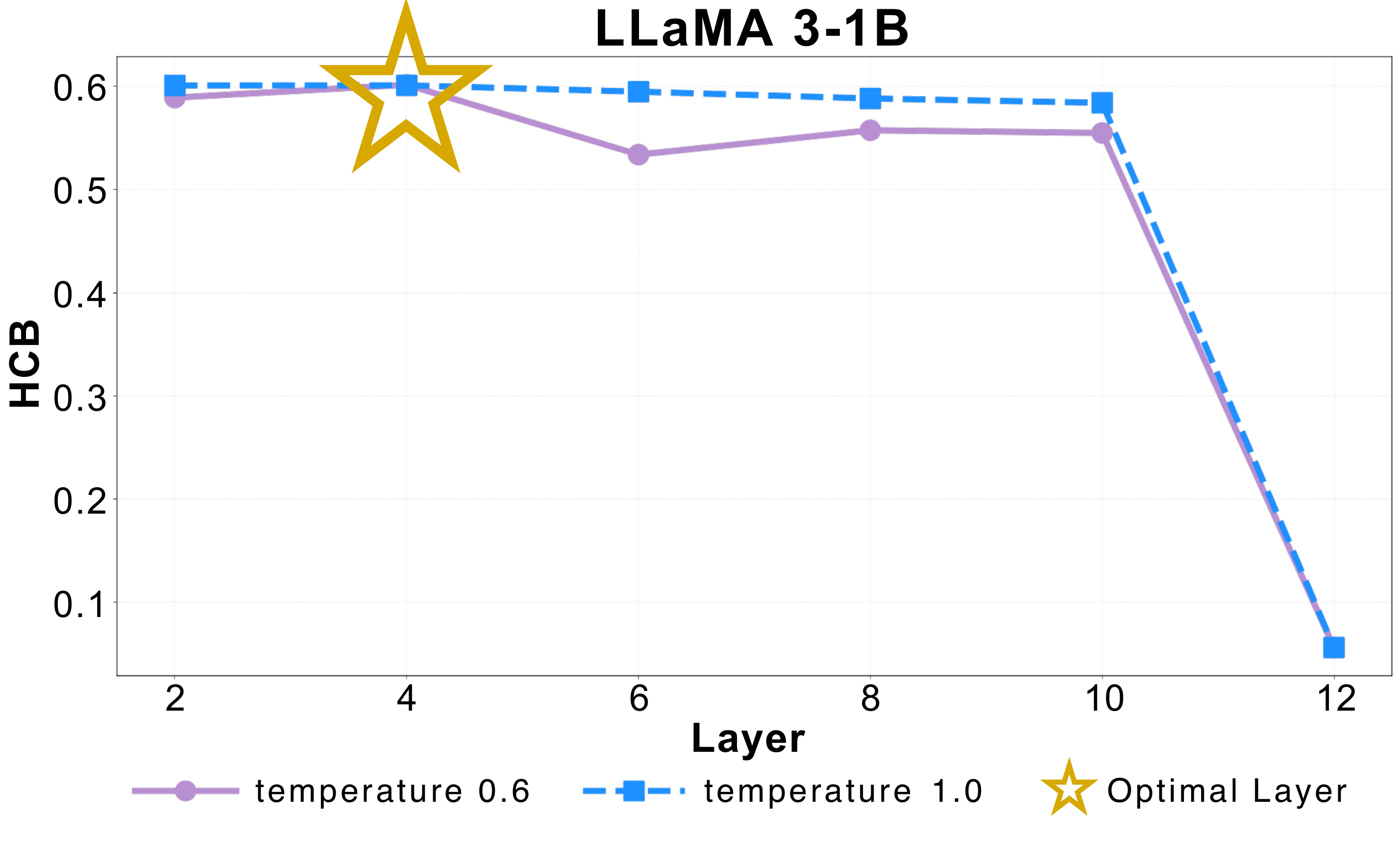}
  \vspace{-1mm}
  \caption{This figure presents the HCB score of the LLaMA3.2-1B. It is evident from the figure that \textbf{\textit{layer-4}} consistently achieves the highest HCB score, regardless of the temperature setting. }
  \vspace{-2mm}
  \label{fig:1b}
\end{figure}

\begin{figure}[t]
  \centering
  \includegraphics[width=0.48\textwidth]{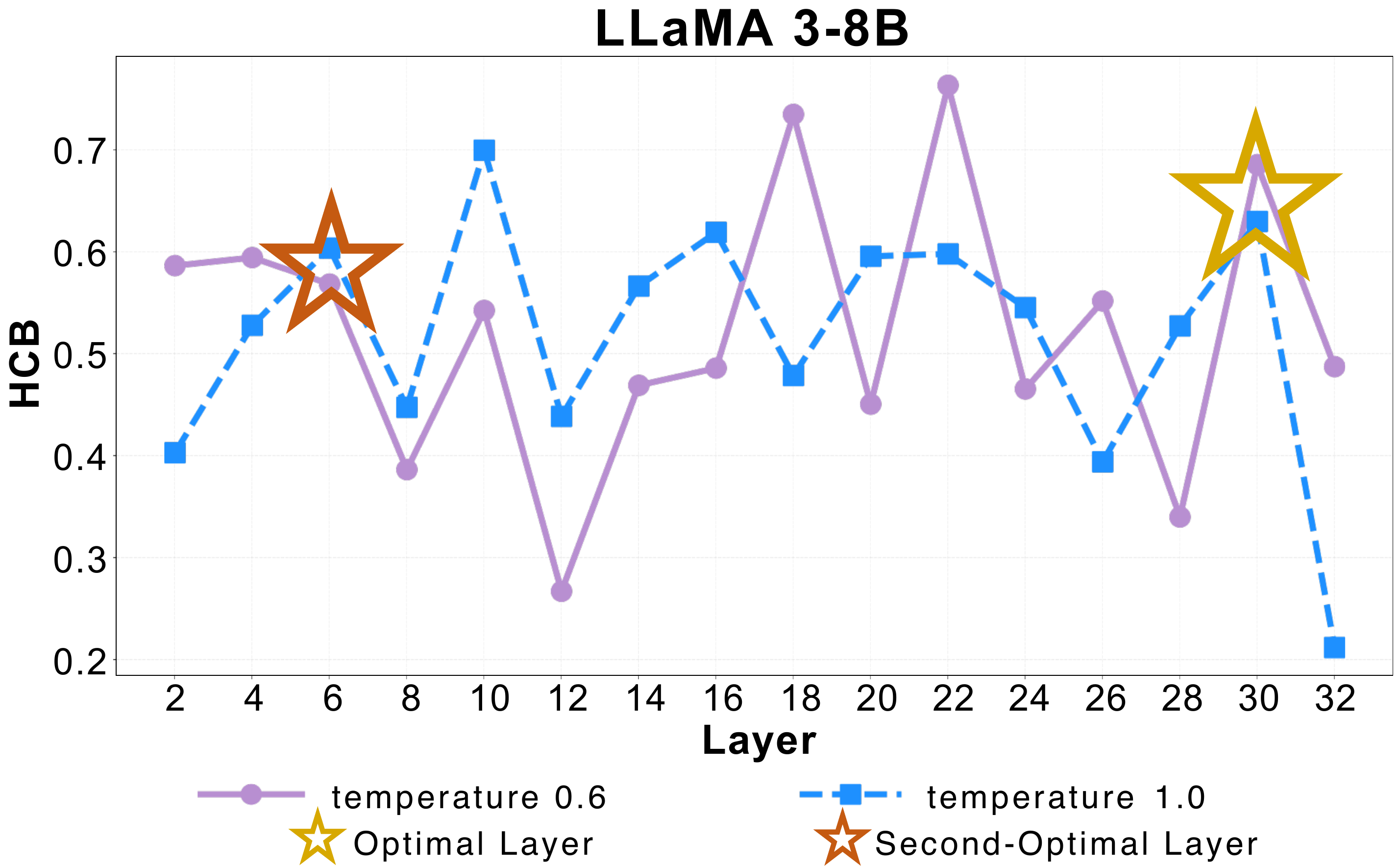}
  \vspace{-1mm}
  \caption{This figure shows the HCB score for LLaMA 3-8B. Although the results indicate \textbf{\textit{layer-30}} is the optimal layer, we further choose \textbf{\textit{layer-8}} to early exit considering the deeper layer causes lower efficiency.}
  
  \vspace{-4mm}
  \label{fig:8b}
\end{figure}

\subsection{Investigate an Optimal Decoding Layer for Early Exit}
In this part, we aim to answer whether there is an optimal decoding layer that achieves the best trade-off between creativity and hallucination, as quantified by our HCB metric. Although conventional approaches typically rely on the final layer’s output, our findings suggest that earlier layers are more likely to produce responses that better balance hallucination and  creativity. By skipping the later layers and selecting outputs from these relatively optimal layers, models can not only be more efficient, but also achieve an optimal balance between hallucination and creativity during generation.

\begin{figure}[t]
  \centering
  \includegraphics[width=0.48\textwidth]{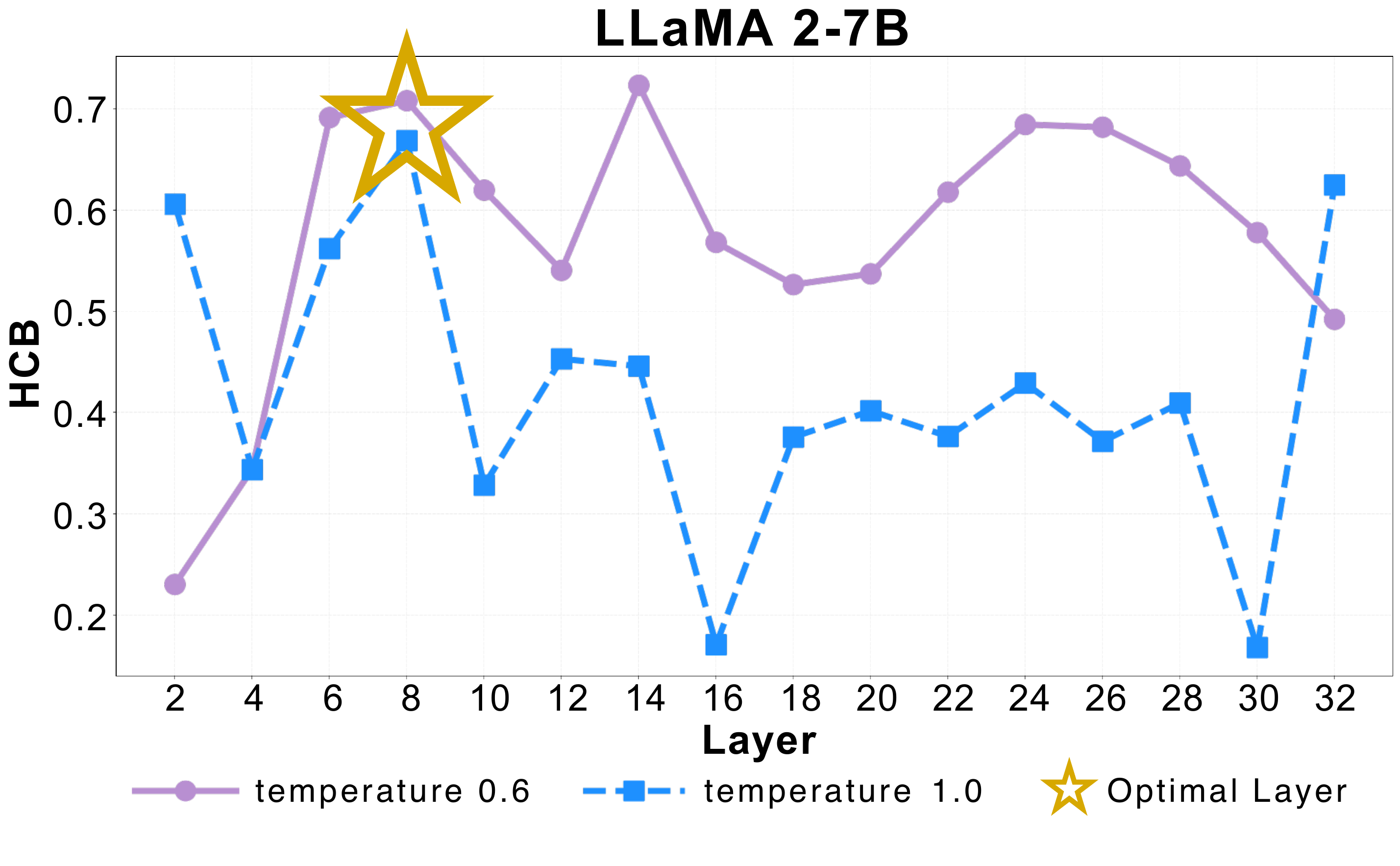}
  \caption{
  This figure illustrates the HCB score of the LLaMA-7B model across its layers. From the results, we can observe that \textbf{\textit{layer-8}} emerges as the optimal layer, whether it is temperature 0.6 or 1.0.}

  \label{fig:7b}
\end{figure}

\begin{figure}[t]
  \centering
  \includegraphics[width=0.48\textwidth]{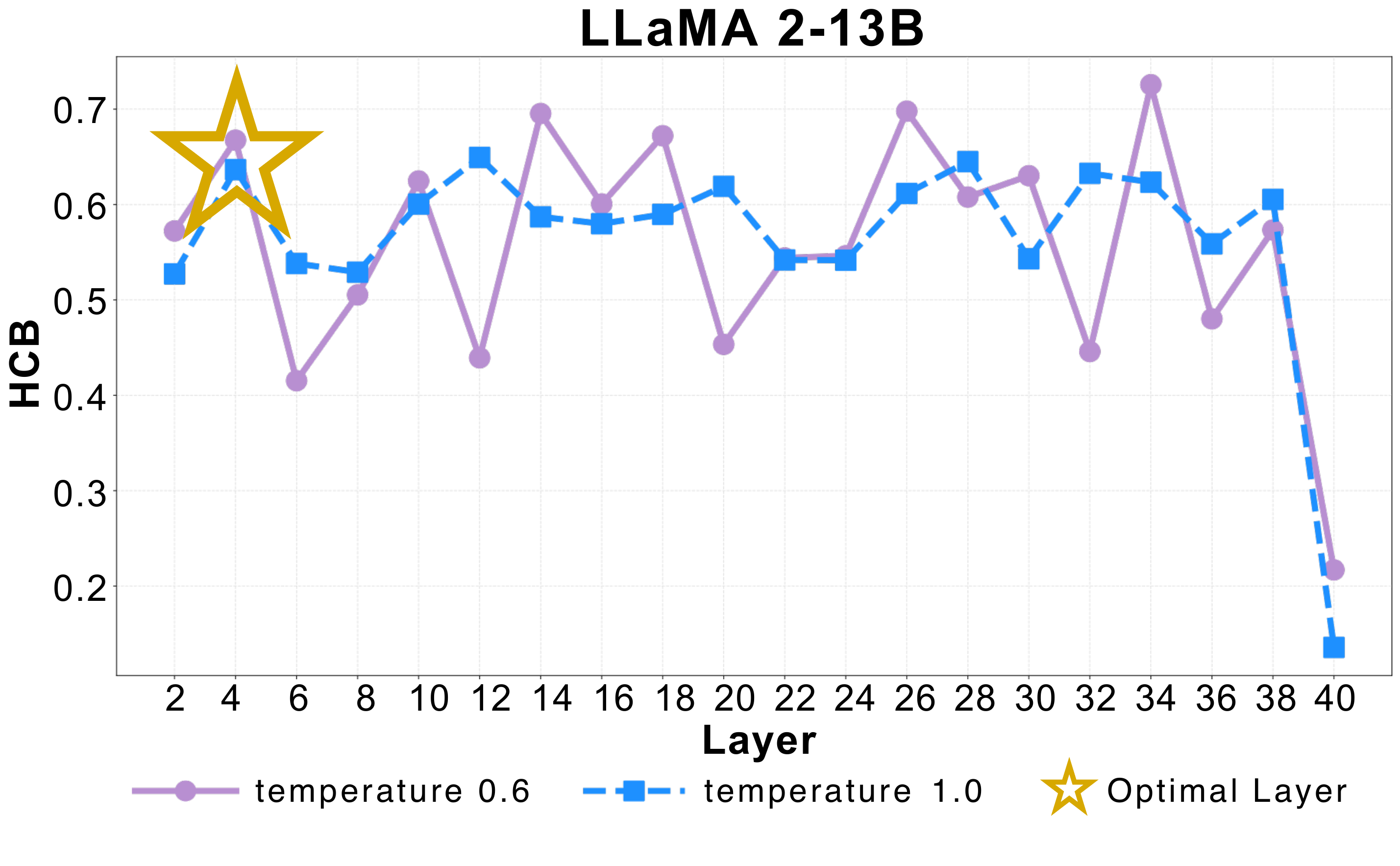}
  \caption{This figure displays the HCB score of the LLaMA-13B model. The results suggest that \textbf{\textit{layer-4}} is the optimal layer since it remains nearly optimal when the temperature changes.}

  \label{fig:13b}
\end{figure}

\paragraph{The output from the final layer is not necessarily the best from a creativity perspective.}
Another key finding from our HCB framework is that final layers, i.e., \textit{layer-12} of LLaMA 3.2-1B, \textit{layer-32} of LLaMA 2-7B, and \textit{layer-40} of LLaMA 2-13B, do not always generate the most creative responses.
While the final layers refine the model’s predictions and improve factual consistency, they often restrict generative flexibility, leading to more deterministic and conservative outputs. In contrast, responses extracted from mid-depth layers tend to exhibit greater creative variation while still maintaining a certain level of factual coherence.
As the results shown in Figure~\ref{fig:1b},~\ref{fig:8b},~\ref{fig:7b},~\ref{fig:13b}, final layer optimization is not necessarily the best strategy and does not always yield superior performance, particularly in applications that prioritize novelty and diversity over absolute factual correctness. Traditional decoding strategies often assume that final layers generate superior responses, but this assumption may need to be revisited and adjusted to better accommodate creative tasks such as storytelling, poetry, and open-ended dialogue generation.

\begin{figure}[t]
  \centering
  \includegraphics[width=0.48\textwidth]{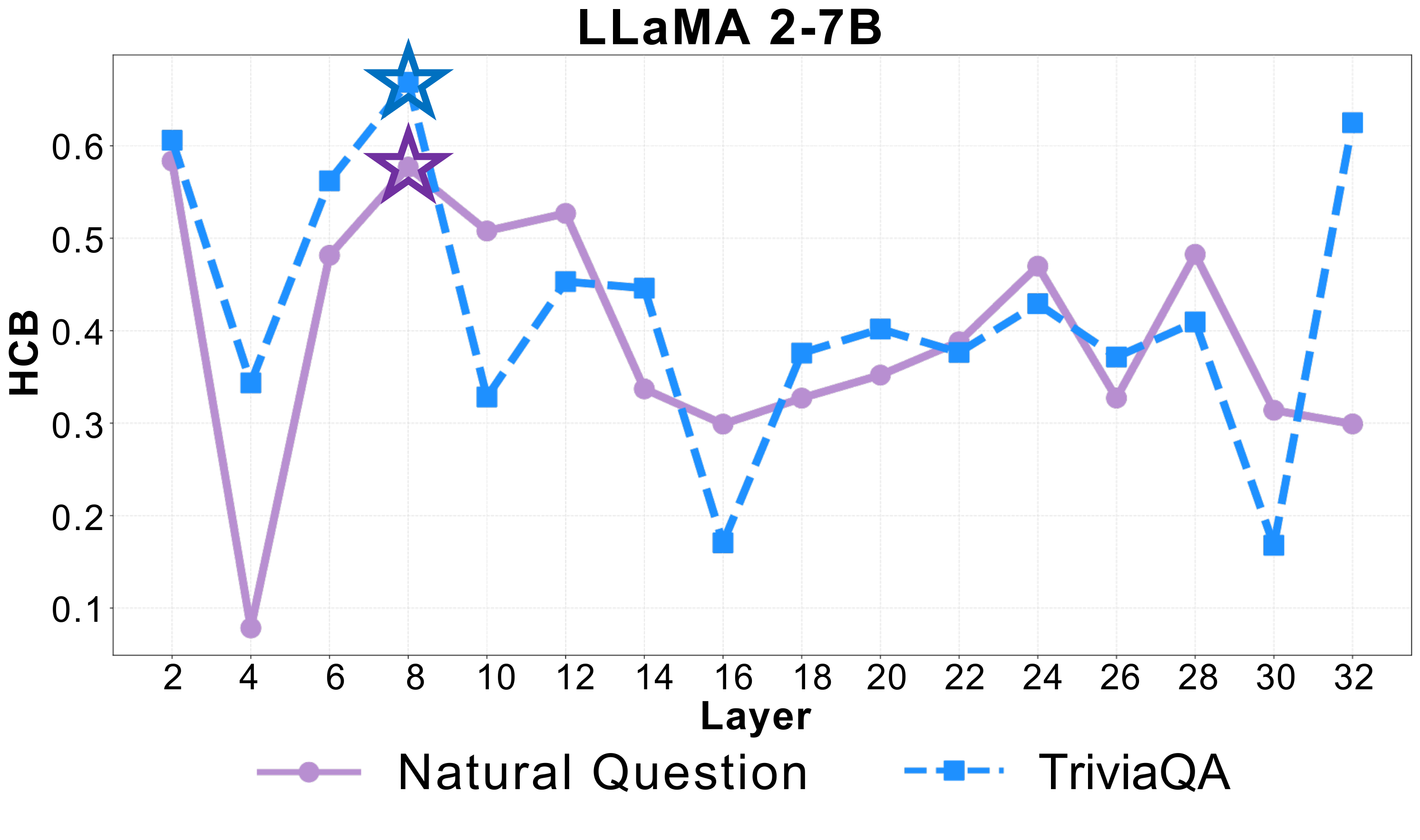}
  \vspace{-7mm}
  \caption{Illustration of the HCB score conducted on LLaMA-7B model at t = 1.0 on TriviaQA and NQ datasets. The results indicate that \textbf{\textit{layer-8}} consistently emerges as the optimal layer for balancing creativity and hallucination in LLMs across both datasets.}
  \vspace{-2mm}
  \label{fig:compare_data}
\end{figure}

\begin{figure}[t]
  \centering
  \includegraphics[width=0.48\textwidth]{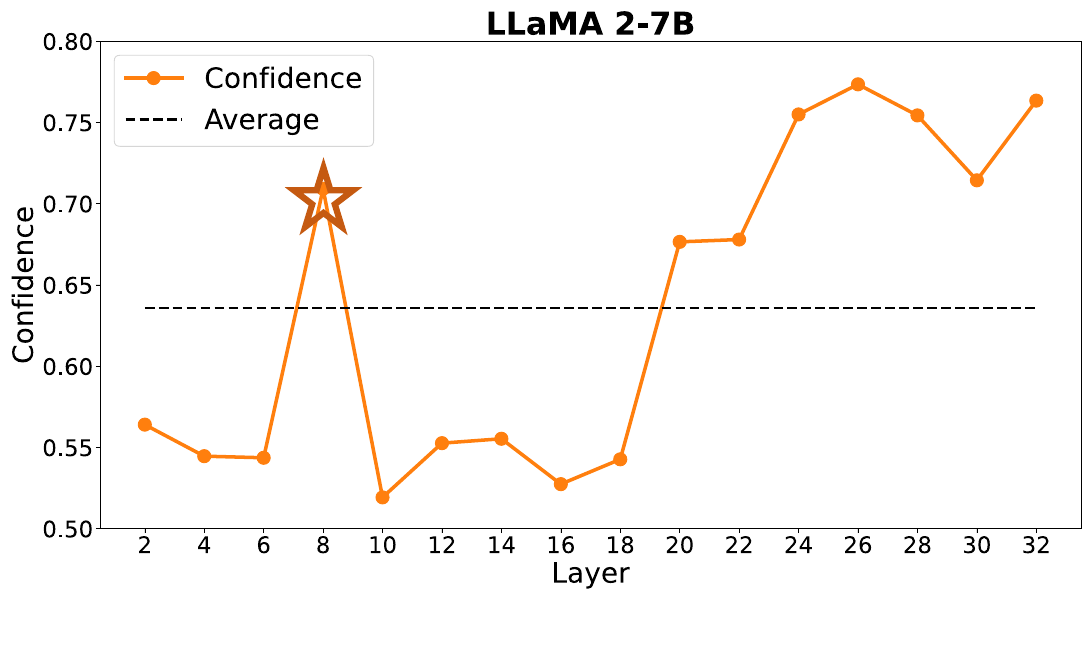}
  \vspace{-6mm}
  \caption{This figure illustrates the variations of confidence across different layers of LLaMA-7B on the TriviaQA dataset. Although the early layers show generally low confidence, there is a sharp peak at \textbf{\textit{layer-8}}, demonstrating our selection on the optimal layer.}
  \vspace{-2mm}
  \label{fig:confidence_7b}
\end{figure}



\paragraph{The optimal layer remains consistently effective across different temperatures and datasets, though it is not always the absolute best choice.} 


Interestingly, our analysis reveals that each model typically has an optimal layer that maintains a stable performance under both temperature 0.6 and 1.0. For instance, in LLaMA 2-7B, \textit{layer-8} consistently balances creativity and factual accuracy across different tasks and temperature settings, despite not being the highest-scoring layer at temperature 0.6. In LLaMA 2-13B, \textit{layer-4} exhibits a stable trade-off between creativity and factual precision. Although \textit{layer-30} is identified as the optimal layer in LLaMA 3-8B, its relatively deep position raises efficiency concerns. Considering the preference for earlier-layer outputs, we suggest \textit{layer-6} for early exit.

It is worth noting that beyond temperature variations, we further analyzed the performance of LLaMA 2-7B on the TriviaQA and NQ datasets, as illustrated in Figure \ref{fig:compare_data}. 
The results demonstrate that the optimal layer in terms of the HCB metric remains consistent across different QA datasets, i.e., \textit{layer-8} remains the one that optimally balances the tradeoff between hallucination and creativity in LLMs. 
The pattern shown in Figure \ref{fig:confidence_7b} further supports the idea that \textit{layer-8} is a key decision-making layer in the model.
This further demonstrates that the identified optimal layer is not only specific to a given model but also has broader generalizability across common QA datasets, verifying the robustness of our HCB-based selection.

\section{Conclusion}
\label{sec:con}


This paper reviews the development of hallucination and creativity in LLMs and proposes a hierarchical evaluation framework, HCL, to explore their interaction across different layers. Additionally, we identify the optimal layer that best balances the tradeoff between hallucination and creativity in LLMs. We have conducted extensive experiments to find key factors influencing both aspects. This study provides a quantitative definition of creativity and offers valuable insights for further exploration of LLM performance across different tasks.

\section*{Ethics Statement}
Our proposed method aims to improve the reliability and creative capabilities of LLMs by analyzing and utilizing responses from different decoding layers. While HCL has the potential to reduce hallucinations while preserving creativity, it is essential to acknowledge the ethical implications associated with our work from the following aspects:

\begin{itemize}
    \item \textbf{Misinformation \& Reliability:}  
    LLMs can generate highly plausible yet incorrect information. By investigating hallucination mechanisms, our study provides insights into distinguishing between factual and misleading outputs. However, our method does not entirely eliminate hallucinations, and caution should be exercised when applying it in high-stakes scenarios such as healthcare or finance.

    \item \textbf{Bias \& Fairness:}  
    LLMs may inherit biases related to gender, ethnicity, and other social factors. Since our framework evaluates hallucination and creativity within existing models, it does not explicitly mitigate bias. Future research should consider fairness-aware approaches to ensure responsible AI deployment.

    \item \textbf{Computational Impact \& Efficiency:}  
    Our layer-wise analysis and early exit strategies aim to optimize computational efficiency, potentially reducing energy consumption in large-scale model inference. However, running extensive experiments with multiple models still requires substantial computational resources.
\end{itemize}
\section*{Limitations}
The correct answer types provided by existing datasets are limited to evaluate the creativity of the LLM's generations.
In addition, our framework is limited to the closed-ended question-answering domain, where a question has multiple objective ground-truth answers so that we can justify the correctness of LLM generated answer. Extensive analysis of HCL on open-ended question-answering tasks in real world scenarios is beyond the scope of the current study and is left as future work. 

The current definition of creativity is relatively narrow, as it distinguishes diversity based on correctness but does not fully consider novelty and originality in subjective or open-ended tasks. In future work, we will expand the evaluation dimensions of creativity to encompass a broader range of creative expressions.  

Additionally, our experiments are limited to a subset of models and do not comprehensively cover LLMs of different scales. In the future, we plan to incorporate \textbf{LLaMA 70B}\citep{touvron2023LLaMA}, \textbf{DeepSeek-R1}\citep{guo2025deepseek}, and \textbf{GPT-4o}\citep{hurst2024gpt}, among other large-scale models, to further validate the applicability of the HCL framework across different model architectures and sizes.


\bibliography{acl_latex}
\clearpage

\appendix

\section{Datasets Statistics.}
\label{app:details:data}
We introduce the two open-domain question answering (QA) datasets used in our study. These datasets are widely employed in QA research and provide a diverse set of real-world questions with multiple valid answers, making them suitable benchmarks for evaluating LLMs in terms of information retrieval, factual accuracy, and creative generation.
\begin{itemize}
    \item \textbf{TriviaQA} \citep{lewis2020retrieval}: 
    TriviaQA is a general knowledge QA dataset that spans multiple domains, including history, science, literature, sports, and entertainment. One of its key characteristics is that each question typically has multiple acceptable correct answers. This diversity makes TriviaQA particularly suitable for evaluating both the correctness and creativity of LLMs. Even in cases where LLMs generate different yet reasonable answers, this dataset allows us to assess their ability to produce factually accurate and contextually diverse responses.  
    In our experiments, we randomly selected 600 samples from TriviaQA, ensuring that each selected question has at least three correct answers.

    \item \textbf{Natural Question} \citep{kwiatkowski2019natural}:  
    Natural Questions (NQ) is a large-scale open-domain QA dataset released by Google, primarily designed for information retrieval and factual question answering. The questions in NQ are sourced from real user queries on Google Search, with corresponding answers typically extracted from Wikipedia pages. Compared to TriviaQA, NQ places a greater emphasis on factual consistency. However, in NQ 2.0, the dataset format evolved from multiple-choice questions to open-ended text generation, providing more flexibility in response formulation. Additionally, many questions in NQ 2.0 now include multiple valid answers, increasing the dataset's adaptability for assessing answer diversity.  
    In our study, we selected 256 questions from the NQ-Open subset, ensuring that each question has at least three correct answers.
\end{itemize}
\paragraph{Model Specifications}  
We conduct experiments using the following LLMs: LLaMA 3-8B, LLaMA 2-7B, LLaMA 2-13B, and LLaMA 3.2-1B, where the numbers indicate the parameter count in billions (B). What's more, we spend average 1066 GPU hours for each model.

\section{Details of LLMs Setups}
\label{sec:Setups}
\paragraph{Temperature}
\label{sec:temperature}
Previous studies have shown that increasing the temperature parameter slightly enhances the novelty of outputs generated by LLMs \citep{peeperkorn2024temperature}. To systematically investigate how temperature influences the trade-off between creativity and hallucination, we set two different temperature values (\( t = 0.6 \) and \( t = 1.0 \)) in our experiments. By comparing the model's performance across different layers under these temperature settings, we aim to examine how temperature affects the model's creative expression while also evaluating its potential impact on hallucination.
\paragraph{Other Hyperparameters}
For all LLMs, the max length of each generation is set to 50 tokens. Besides, all other parameters remain consistent with Layer-Skip.
For our evaluation framework, we set the sampling time to 50 to ensure there are enough response evaluations. During the HCB score calculation, we define the formula as follows:
\[
S_{HCB}^i = w_c \times S_C^i \;+\; w_h \times \bigl(1 - S_H^i\bigr),
\]

where both of \(w_c\) and \(w_h\) are set to 0.5.
\section{Details of semantic cluster}

\begin{enumerate}
    \item \textbf{Answer Embedding}: For each correct answer $a$, we compute a dense vector representation $\vec{v}_a$:
    \[
    \vec{v}_a = \mathrm{Encoder}(a),
    \]
    where $\mathrm{Encoder}$ is the SentenceTransformer model capturing contextual and semantic information.

    \item \textbf{Cosine Similarity}: We calculate the cosine similarity between $\vec{v}_a$ and each vector $\vec{v}_u$ in the set of previously identified unique answers:
    \[
    \mathrm{sim}(\vec{v}_a, \vec{v}_u) = \frac{\vec{v}_a \cdot \vec{v}_u}{\|\vec{v}_a\|\|\vec{v}_u\|}.
    \]
    The similarity ranges from $-1$ to $1$, with higher scores indicating stronger semantic resemblance.

    \item \textbf{Thresholding}: If $\mathrm{sim}(\vec{v}_a, \vec{v}_u) \ge \tau$ (we set $\tau=0.8$), then $a$ is considered semantically equivalent to an existing unique answer. Otherwise, $a$ is added to the set of unique answers. This threshold avoids over-clustering or splitting near-identical answers.
\end{enumerate}

\section{Layer-wise Confidence Measurement}
We adopt P(True) \citep{kadavath2022language} to measure the confidence of each decoding layer of the LLM on its generations.
Specifically, we follow \citep{kadavath2022language} and prompt the LLM layer by layer to judge whether its own generated answer is correct. Our prompt followed the following template:

\begin{tcolorbox}[
    colframe=black,  
    colback=white,   
    rounded corners=north, 
    arc=1mm, 
    boxrule=.4mm, 
    width=\linewidth, 
    coltitle=black, 
    colbacktitle=white,
    title={\textbf{P(True)}}, 
    fonttitle=\bfseries, 
]

\textbf{Question}: [Question]

\textbf{Possible Answer}: [LLM Answer]

\vspace{5pt}

Is the possible answer:\\
(A) False\\
(B) True

\vspace{5pt}

\textbf{The possible answer is}: 

\end{tcolorbox}

\end{document}